# Breast Cancer Classification from Histopathological Images with Inception Recurrent Residual Convolutional Neural Network

Md Zahangir Alom, Chris Yakopcic, Tarek M. Taha, and Vijayan K. Asari
Department of Electrical and Computer Engineering, University of Dayton, OH, USA
Emails: {alomm1, cyakopcic1, ttaha1, vasari1}@udayton.edu

Abstract

The Deep Convolutional Neural Network (DCNN) is one of the most powerful and successful deep learning approaches. DCNNs have already provided superior performance in different modalities of medical imaging including breast cancer classification, segmentation, and detection. Breast cancer is one of the most common and dangerous cancers impacting women worldwide. In this paper, we have proposed a method for breast cancer classification with the Inception Recurrent Residual Convolutional Neural Network (IRRCNN) model. The IRRCNN is a powerful DCNN model that combines the strength of the Inception Network (Inception-v4), the Residual Network (ResNet), and the Recurrent Convolutional Neural Network (RCNN). The IRRCNN shows superior performance against equivalent Inception Networks, Residual Networks, and RCNNs for object recognition tasks. In this paper, the IRRCNN approach is applied for breast cancer classification on two publicly available datasets including BreakHis and Breast Cancer Classification Challenge 2015. The experimental results are compared against the existing machine learning and deep learning-based approaches with respect to image-based, patch-based, image-level, and patient-level classification. The IRRCNN model provides superior classification performance in terms of sensitivity, Area Under the Curve (AUC), the ROC curve, and global accuracy compared to existing approaches for both datasets.

## Introduction

Nowadays, cancer is one of the leading causes of morbidity and mortality around the world. Approximately 14.5 million people have died due to cancer, and it is estimated that this number will be above 28 million by 2030. According to a study by the American Cancer Society (ACS), in the USA the estimated deaths due to breast cancer account for approximately 14% of all cancer deaths (a total of 41,000 in 2017) which is in the second-leading cause of cancer death in women after lung and bronchus cancer. Additionally, breast cancer accounts for 30% of all newly discovered cancer cases. Breast cancer is the most frequently diagnosed cancer in women in the USA. A biopsy followed by microscopic image analysis is common when diagnosing breast cancer [1]. A breast tissue biopsy allows the pathologist to histologically access the microscopic level structures and components of the breast tissue. These histological images allow to the pathologist to distinguish between the normal tissue, non-malignant (benign) tissue, and malignant lesions. The resulting information is then used to perform a prognostic evaluation [2].

Benign lesions refer changes in normal tissue of breast parenchyma, and are not related to the progression of malignancy. There are two different carcinoma tissue types including in-situ and invasive. The in-situ tissue type refers to tissue contained inside the mammary ductal-lobular. On the other hand, the invasive carcinoma cells spread beyond the mammary ductal-lobular structure. The tissue samples that are collected during biopsy are commonly stained with Hematoxylin and Eosin (H&E) prior to the visual analysis performed by the specialist. During the diagnosis process, the affected region is determined from whole-slide tissue scans [3]. In addition, the pathologist analyzes microscopic images of the tissue samples from the biopsy with different magnification factors. Nowadays, to produce the correct diagnosis, the pathologist considers different characteristics within the images including patterns, textures, and different morphological properties [4]. Analyzing images with different magnification factors requires panning, zooming, focusing, and scanning of each image in its entirety. This process is very time consuming and tiresome, as a result this manual process sometimes leads to inaccurate diagnosis for breast cancer identification. Due to the advancement of digital imaging techniques in the last decade, different computer vision and machine learning techniques have been applied for analyzing the pathological images at a microscopic resolution [4,5]. These approaches could help to automate some of the tasks related to the pathological workflow in diagnosis system. However, an efficient and robust image processing algorithm is necessary for use in clinical practices. Unfortunately, traditional approaches are unable to fulfill the expectation. As a result, we are still far from practical application of automatic breast cancer detection based on histological images [5].

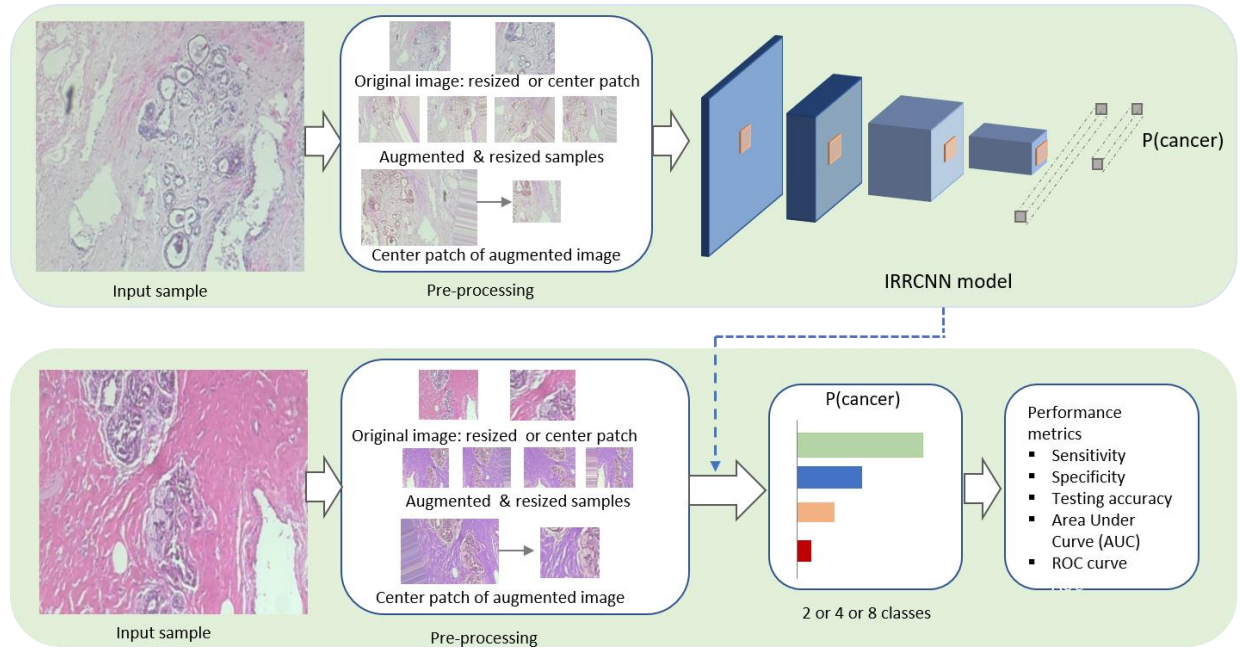

Figure 1. Implementation diagram for breast cancer recognition using the IRRCNN model. The upper part of this figure shows the steps that are used for training the system, and the lower part of this figure displays the testing phase where the trained model is used. These results are evaluated with a number of different performance metrics.

However, recent developments in Deep Learning (DL) have already shown vast potential with state-of-the-art performance on different recognition tasks in the field of computer vision and image processing, speech recognition, and natural language understanding [6]. These approaches have been applied in different modalities of medical imaging including pathological imaging with superior performance in classification, segmentation, and detection [7]. In some cases, the DL based systems have become part of the workflow for clinical practices with pathologists and doctors. Some examples include dermatologist-level performance for skin cancer detection, diabetic retinopathy, neuroimaging for analysis of brain tumors and Alzheimer disease, lung cancer detection, and breast cancer detection and classification [7]. Although these approaches have shown tremendous success in medical imaging, they require very large amount of label data, which is still not available in this domain of applications for several reasons. Most significantly, it requires a lot of expertise to annotate a dataset which is very expensive. In this paper, we propose a DL based approach for breast cancer recognition system using the IRRCNN model which is evaluated using the BreaKHis and Breast Cancer Classification Challenge 2015 datasets. The contributions of this paper are summarized as follows:

1. Successful magnification factor invariant binary and multi-class breast cancer classification using the IRRCNN model.
2. Experiments have been conducted on recently released publicly available datasets for breast cancer histopathology (such as the BreaKHis dataset) where we evaluated image and patient level data with different magnifying factors (including 40×, 100×, 200×, and 400×).
3. Image-based and patch-based evaluation was performed for both the BreaKHis and Breast Cancer Classification Challenge 2015 datasets

4. The experimental results are compared against recently proposed deep learning and machine learning approaches, and our proposed model provides superior performance when compared to the existing algorithms for breast cancer classification.

This paper is organized as follows: Section II discusses related work. The architectures of the IRRCNN model are presented in Section III. Section IV explains the datasets, experiments, and results. Finally, the conclusion and future direction are discussed in Section V.

# Related Works

Significant effort has been put forth for breast cancer (BC) recognition from histological images in the last decade, where most efforts are made to classify the two fundamental types of breast cancer (benign and malignant) using Computer Aided Diagnosis (CAD). Before the deep learning revolution, machine learning approaches including the Support Vector Machine (SVM), Principle Component Analysis (PCA), and Random Forest (RF) were used to study data whose features were extracted with Scale Invariant Feature Extraction (SIFT), Local Binary Patterning (LBP), Local Phase Quantization (LPQ), the Gray-Level Co-occurrence Matrix (GLCM), Threshold Adjacency Statistics (TAS), and Parameter Free TAS (PFTAS). In 2016, one of the very popular databases for BC classification problem was released, and one research group reported approximately 85.1% accuracy utilizing SVM and PFTAS features for patient-level analysis [8], which was the highest recognition accuracy at the time. Another work was published in 2013 where different algorithms (including K-means, fuzzy C-means, competitive learning neural networks, and Gaussian mixture models) were used for nuclei classification on a dataset with 500 samples from 50 patients. The accuracies were reported for binary classes (benign versus malignant). This work produced accuracies ranging from 96 – 100 percent [9].

A machine learning system for breast cancer recognition based on Neural Networks (NN) and SVM was published in 2013 that reported 94% recognition accuracy on a dataset that consisting of 92 samples [10]. Another method was proposed based on cascading with a rejection option that was tested on a dataset with 361 samples from the Israel Institute of Technology, and it reported around 97% classification accuracy [11]. For the most part, research in this area has been conducted using a very small number of samples from primarily private datasets. Recently a survey was published on histological image analysis for breast cancer detection and classification that clearly describes the dualities and limitations of different publicly available annotated datasets [12]. An effective framework has been proposed with color texture features and multiple classifiers utilizing voting technique that reported approximately 87.53% average recognition rate for patient level BC classification. In this implementation, the SVM, the Decision Tree (DT), a Nearest Neighbor Classifier (NNC), Discriminant Analysis (DA), and Ensemble classifiers were used. Before 2017, this system achieved the best recognition accuracy of all machine learning based approaches [13].

Furthermore, many works have already been published that discuss breast cancer recognition using DL approaches, where CNN variants are applied for classification. A few of these experiments are conducted with the BreaKHis dataset. In 2016, a magnification independent breast cancer classification was proposed based on a CNN where different sized convolution kernels (7×7, 5×5, and 3×3) were used. They performed patient level classification of breast cancer with CNN and multi-task CNN (MTCNN) models and reported an 83.25% recognition rate [14]. In the same year, another work was published based on a model similar to AlexNet with different fusion techniques (including sum, product, and max) for image and patient level classification of breast cancer. This paper reports 90% and 85.6% average recognition accuracy with the max fusion method for images and patient level classification respectively [15]. Another deep learning-based method was published in 2017. In this work, a pretrained CNN was used to extract the feature vectors, and eventually the feature vectors were used as the input to a classifier. This method was called DeCAF and achieved a recognition accuracy of 86.3% and 84.2% at the patient level and image level respectively [16].

The CNN model was used for the classification of H&E stained breast biopsy images from another challenging dataset in 2017 [17]. The images were classified according to four different classes: normal tissue, benign lesion, in-situ carcinoma, and invasive carcinoma. Images were also classified in terms of binary classes, carcinoma (which includes normal and benign tissue) and non-carcinoma (which includes the in-situ and invasive carcinoma classes) are considered. Work in [17] provides results for both image-based and patch-based evaluation. The CNN based approach achieved approximately 77.8% recognition accuracy when performing the four-class experiment, and 83.3% recognition accuracy for the binary class experiment when tested with the BC Classification Challenge 2015 dataset. Recently, multi-classification of breast cancer from histopathological images was presented using a structured deep learning model called CSDCNN. This new DL architecture shows superior performance when compared to different machine learning and deep learning-based approaches on the BreaKHis dataset. This model shows state-of-the-art performance for both image-level and patient-level classification. An average of 93.2% accuracy for patient-level breast cancer classification has been reported [18]. In 2017, different SMV based techniques were applied for breast cancer recognition, an accuracy of 94.97% for data with a 40× magnification factor was achieved using an Adaptive Sparse SVM (ASSVM) [28]. However, our work presents an application of a new deep learning model called the Inception Recurrent Residual Convolutional Neural Network (IRRCNN) for BC classification on both the BreaKHis and 2015 Breast Cancer Classification Challenge datasets.

## IRRCNN Model for Breast Cancer Recognition

DL approaches show tremendous success in cases where sufficient labeled data is available, and several advanced deep learning approaches have been proposed that have shown state-of-the-art performance in different modalities of computer vision and medical imaging in the last few years [6,7]. The Inception Recurrent Residual Convolutional Neural Network (IRRCNN) [19,20] is one out of many which is an improved hybrid DCNN architecture based on inception [21], residual networks [23], and the RCNN architecture [24]. The main advantage of this model is that it provides better recognition performance using the same number or fewer network parameters when compared to alternative equivalent deep learning approaches including inception, the RCNN, and the residual network. In this model, the inception-residual units are utilized with respect to the Inception-v4 model [2]. The IRRCNN has been compared against equivalent inception-residual networks, and it shows better performances [19]. The IRRCNN model is comprised of stacks that include both inception recurrent residual units (IRRU) and transition units. The entire model is shown in Fig 1. The overall model consists of several convolution layers, IRRUs, transition blocks, and a softmax at the output layer. A pictorial view of the IRRU is shown in Fig. 2.

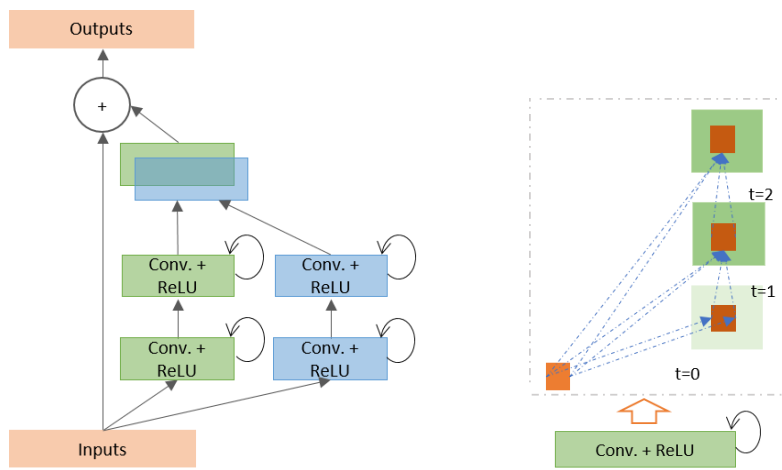

Figure 2. Diagrams displaying the Inception Recurrent Residual Unit (IRRU) consisting of the inception unit and recurrent convolutional layers that are merged by concatenation, and the residual units (summation of the input features with the outputs of the inception unit can be seen just before the output block).

The most important unit in the IRRCNN architecture is the IRRU, which includes Recurrent Convolutional Layers (RCLs), inception units, and a residual layer. The inputs are fed into the input layer, then passed through inception units where RCLs are applied, and finally the outputs of the inception units are added to the inputs of the IRRU. The recurrent convolution operations are performed with respect to the different sized kernels in the inception unit. Due to the recurrent structure within the convolution layer, the outputs at the present time step are added to the outputs of previous time step. The outputs at the present time step are then used as inputs for the next time step. The same operations are performed with respect to the time steps that are considered. For example, here *t = 2 (0~2)* means that one feed forward convolution along with 2 RCLs are included in IRRU. The operation of the RCLs with respect to the different time steps (*t* = 2 *(0~2)* and *t* = 3 *(0~3)*) is shown in Fig. 2. Due to the residual connectivity in the IRRU, the input and output dimensions do not change. The IRRU simply performs an accumulation of feature maps with respect to the time steps. Thus, better feature representation is ensured, and this system achieves superior recognition accuracy with the same number of network parameters.

The operations of the RCL are performed with respect to discrete time steps that are expressed according to the IRRCNN in [19]. Let's consider the $x_l$ input sample in the $l^{th}$ layer of the IRRCNN block, and the unit $(i,j)$ from an input sample in the $k^{th}$ feature map in the RCL. Additionally, let's assume the output of the network $O_{ijk}^l(t)$ is at time step *t*. Given this information, the output can be expressed as in equation (1).

$$O_{ijk}^l(t) = \left(w_k^f\right)^T * x_l^{f(i,j)}(t) + (w_k^r)^T * x_l^{r(i,j)}(t-1) + b_k \tag{1}$$

Here $x_l^{f(i,j)}(t)$ and $x_l^{r(i,j)}(t-1)$ are the inputs for the standard convolution layers and for the $l^{th}$ RCL respectively. The $w_k^f$ and $w_k^r$ values are the weights for the standard convolutional layer and the RCL of the $k^{th}$ feature map respectively, and $b_k$ is the bias.

$$y = f(O_{ijk}^l(t)) = \max(0, O_{ijk}^l(t)) \tag{2}$$

In equation (2), $f$ is the standard Rectified Linear Unit (ReLU) activation function. We have also explored the performance of this model with the Exponential Linear Unit (ELU) activation function in the following experiments. The outputs $y$ of the inception units for the different size kernels and average pooling layer are defined as $y_{1x1}(x)$, $y_{3x3}(x)$, and $y_{1x1}^p(x)$ respectively. The final outputs of Inception Recurrent Convolutional Neural Network (IRCNN) unit are defined as $\mathcal{F}(x_l, w_l)$ which can be expressed as in equation (3).

$$\mathcal{F}(x_l, w_l) = y_{1x1}(x) \odot y(x) \odot y_{1x1}^p(x) \tag{3}$$

Here $\odot$ represents the concatenation operation with respect to the channel or feature map axis. The outputs of the IRCNN unit are then added with the inputs of the IRRCNN block. The residual operation of the IRRCNN block can be expressed as in equation (4).

$$x_{l+1} = x_l + \mathcal{F}(x_l, w_l) \tag{4}$$

In equation (4), $x_{l+1}$ refers to the inputs for the immediate next transition block, $x_l$ represents the input samples of the IRRCNN block, $w_l$ represents the kernel weights of the $l^{th}$ IRRCNN block, and $\mathcal{F}(x_l, w_l)$ represents the outputs from of $l^{th}$ layer of the IRCNN unit. However, the number of feature maps and the dimensions of the feature maps for the residual units are the same as in the IRRCNN unit shown in Figure 2. Batch normalization is applied to the outputs of the IRRU [24]. Eventually, the outputs of this IRRU are fed to the inputs of the immediate next transition unit.

In the **transition unit,** different operations including convolution, pooling, and dropout are performed depending upon the placement of the transition unit in the model. The inception units are included in the transition unit. The down-sampling operations are performed in the transition units, where we perform max-pooling operations with a 3×3 patch and a 2×2 stride. The non-overlapping max-pooling operation has a negative impact on model regularization, therefore we used overlapped max-pooling for regularizing the network which is very important when training a deep network architecture [21]. The late use of a pooling layer helps to increase the non-linearity of the features in the network, as this results in higher dimensional feature maps being passed through the convolution layers in the network. We used two special pooling layers in the model with three IRRCNN units and one transition unit for this implementation.

We used only 1×1 and 3×3 convolution filters in this implementation, as inspired by the NiN [6] and Squeeze Net [6] models. This also helps to keep the number of network parameters at a minimum. The benefit of adding a 1×1 filter is that it helps to increase the non-linearity of the decision function without having any impact on the convolution layer. Since the size of the input and output features does not change in the IRRCNN units, the result is just a linear projection on the same dimension, and non-linearity is added to the RELU and ELU activation functions. We used a 0.5 dropout after each convolution layer in the transition block. Finally, we used a softmax, or normalized exponential function layer at the end of the architecture. For an input sample *x*, a weight vector *W*, and *K* distinct linear functions, the softmax operation can be defined for the $i^{th}$ class as in equation (5).

$$P(y = i|x) = \frac{e^{x^T w_i}}{\sum_{k=1}^{K} e^{x^T w_k}} \tag{5}$$

The proposed IRRCNN model has been investigated through a set of experiments on different benchmark datasets, and the results have been compared across several different models.

The IRRCNN model is evaluated with different numbers of convolutional layers in the convolution blocks, and the number of layers is determined with respect to time step *t*. In these implementations, *t* = 2 refers to a RCL block that contains one forward convolution followed by two RCLs [19]. For both breast cancer recognition datasets, we used a model with two convolutional layers at the beginning, four IRCNN blocks followed by transition blocks, a fully connected layer, and a softmax layer at the end of the model. For this model, we considered 32 and 64 feature maps for the first three convolutional layers, and we used 128, 256, 512, and 1024 feature maps in the first, second, third, and fourth IRRCNN blocks respectively. Batch Normalization (BN) is used in each IRCNN block. This model contains approximately 9.3 million network parameters.

# Experimental Results and Discussion

## Experimental setup

To demonstrate the performance of the IRRCNN models, we have tested them on two different BC datasets: the BreakHis dataset, and the Breast Cancer Classification Challenge 2015 dataset for both binary and multi-class BC classification. The following paragraph discusses both datasets in detail. For this implementation, the Keras [30], and Tensor Flow [31] frameworks were used on a single GPU machine with 56G of RAM and an NIVIDIA GEFORCE GTX-980 Ti. We considered different criterion for pathological image analysis in this implementation. In most cases, the dimensions of the Whole Slide Images (WSI) are larger than typical digital images. In addition, the pathological images are acquired with different magnification factors. In some cases, the image size is larger than 2000 × 2000 pixels. However, in this case the images are typically fed to the model as several patches. There are two common processes used for patch selection, one of which is a random crop method where the patches are cropped from random location within an input sample. The alternative is to use sequential and non-overlapping patches. We considered both methods in this implementation.

Table 1. Statistics for the main and subclass samples and number of patients for the BreaKHis dataset.

| Classes | Subclasses | Number of Patients | Magnification factors | | | | Total |
|---|---|---|---|---|---|---|---|
| | | | 40× | 100× | 200× | 400× | |
| Benign | A | 4 | 114 | 113 | 111 | 106 | 444 |
| | F | 10 | 253 | 260 | 264 | 237 | 1014 |
| | TA | 3 | 109 | 121 | 108 | 115 | 453 |
| | PT | 7 | 149 | 150 | 140 | 130 | 569 |
| Malignant | DC | 38 | 864 | 903 | 896 | 788 | 3451 |
| | LC | 5 | 156 | 170 | 163 | 137 | 626 |
| | MC | 9 | 205 | 222 | 196 | 169 | 792 |
| | PC | 6 | 145 | 142 | 135 | 138 | 560 |
| Total | | 82 | 1995 | 2081 | 2013 | 1820 | 7909 |

## Datasets

**BreakHis:** The BreaKHis dataset is publicly available and is commonly used to study the breast cancer classification problem. This dataset contains 7909 samples each falling within two main classes: benign or malignant. The benign subset contains 2440 samples and the malignant subset contains 5429 samples. The samples are collected from 82 patients with different magnification factors including 40×, 100×, 200×, 400×. Some of the example images with a 400× magnification factor are shown in Figure 3. Each class has four subclass, the four types of benign cancer are Adenosis (A), Fibroadenoma (F), Tubular Adenoma (TA), and Phyllodes Tumor (PT). The four subclasses of malignant cancer are Ductal Carcinoma (DC), Lobular Carcinoma (LC), Mucinous Carcinoma (MC), and Papillary Carcinoma (PC). The statistics for this dataset are given in Table 1. In this experiment, we used 70% of the samples for training and 30% of the samples for testing, per the work in [12,18]. To generalize the classification task to perform successfully when testing new patients, we ensure that the patients selected for training are not used during testing. Per the experimental design in [12], we reported the average accuracy after successfully completing five trials.

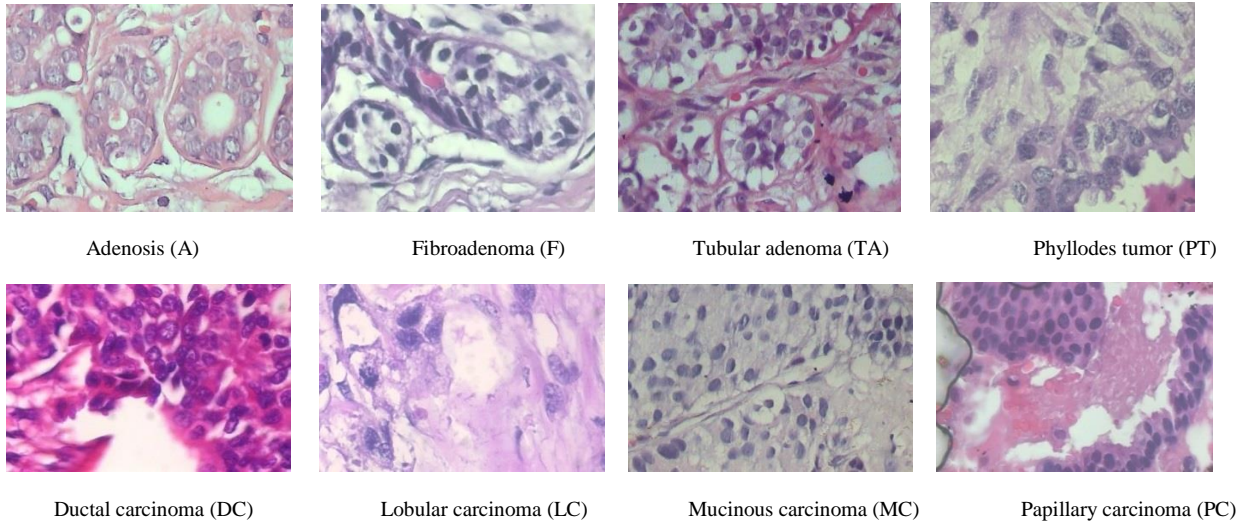

Figure 3. The first row shows the four types of benign tumors, and the second row shows the malignant tumors. The magnification factor of these images is 400×.

For data augmentation, we generated 21 samples from each single input sample with different augmentation techniques including rotation, flipping, shearing, and translation. Therefore, the total number of samples was increased by 21 times. For example, the total number of images available at a 40× magnification is now 41,895. We generated 43701, 42273, and 38220 patches from the original samples for the 100×, 200×, and 400× magnification factors respectively. Thus, total number of augmented samples for all magnification factors is 166,068. We evaluated the image-level and patient-level performance for both binary and multi-class breast cancer recognition.

Table 2. Statistics for the 2015 BC Classification Challenge dataset.

| Methods | Non-Carcinoma | | Carcinoma | | Total |
|---|---|---|---|---|---|
| | Normal | Benign | In situ | Invasive | |
| Image-wise | 55 | 69 | 63 | 62 | 249 |
| Augmented samples | 1155 | 1449 | 1323 | 1302 | 5229 |
| Random patches | 9716 | 12057 | 11059 | 10875 | 43,707 |

**Breast Cancer Classification Challenge 2015:** This dataset consists of very high resolution (2040×1536) digital pathology images, which are annotated H&E stained images for breast cancer classification released in 2015 [17,27]. This dataset contains total 249 samples, from which 229 samples are separated for training, and the remaining samples are considered for testing, per the work in [17]. The images were labeled by two pathologists and the overall context has been considered without specifying the area of interest. Each image is assigned one of the following four categories: (a) normal tissue (b) benign (c) in-situ, and (d) invasive carcinoma. Sample images displaying the four different types of BC are shown in Figure 4. Each class has about 60 samples, which resolves the class imbalance problem for classification tasks. In this implementation, the model is evaluated for binary and multi-class BC classification. In case of the binary classification problem, the normal tissue and benign subsets are considered class one, and the in-situ and invasive carcinoma subsets are considered to be a part of class two. According to a visual analysis of the dataset, it is observed that the nuclei radius ranges from 3 to 11 pixels (or 1.26µm to 4.62 µm). Therefore, patches with size 128×128 pixels are able to cover enough of the tissue structure (in accordance with the experiment conducted in [17]).

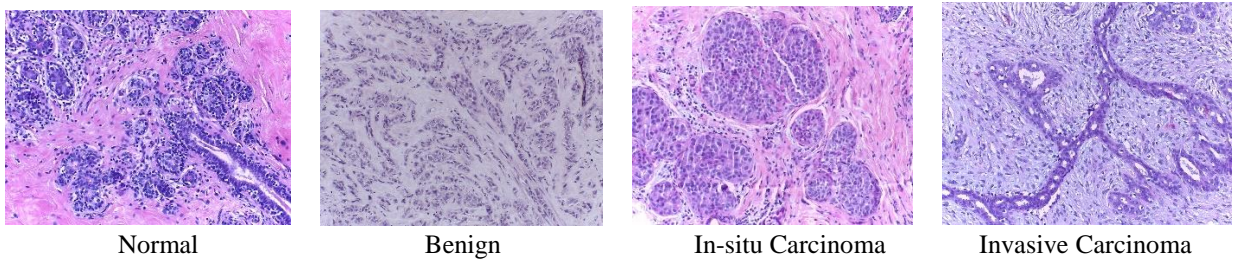

| Normal | Benign | In-situ Carcinoma | Invasive Carcinoma |

Figure 4. Sample images of four types of breast cancer (normal, benign, in situ carcinoma, and invasive carcinoma) from the 2015 BC Classification Challenge dateset.

We have conducted experiments using both image-wise and patch-wise evaluation. For image-wise classification, we used three different approaches: first, we resized the input samples to 128×128 pixels which significantly degrades the information contained in the samples. Second, different data augmentation techniques were applied to the resized images where 20 different augmented samples were generated for each sample. Third, 200 random patches were cropped to create a patch database for training and testing the model. A Winner Take All (WTA) method was used to produce the results where the final class was determined based on the class where the maximum number of patches were nominated. The labels of the patches are considered to have same class label as the original images. On the other hand, using patch-wise approach: first, 128×128 pixel center patches were cropped from an input sample. Second, the augmentation techniques were applied on the center patches and 20 augmented samples per patch were generated. Third, we evaluated the model with 200 randomly selected patches with a size of 128×128 pixels from a single image. The statistics for the image-wise and patch-wise approach are given in Table 2.

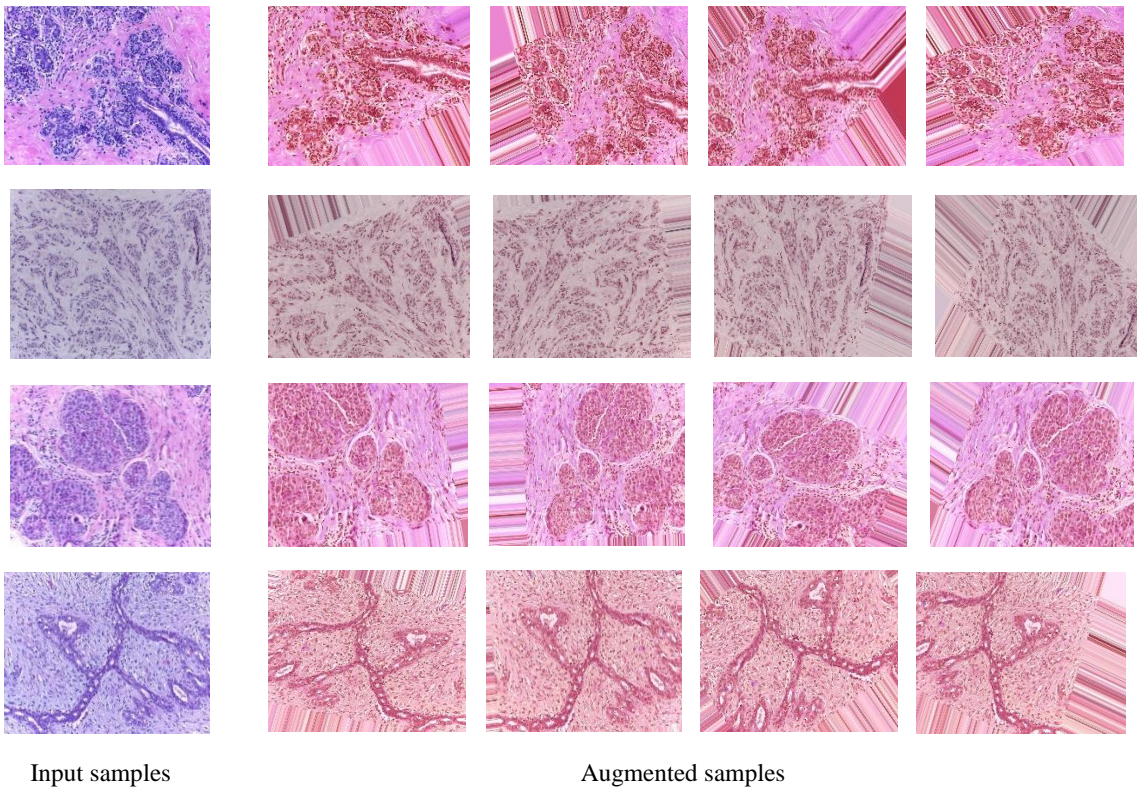

Input samples　　　　　　　　　　　Augmented samples

Figure 5. Four example images with corresponding augmented images. The actual images are shown on the left, and four augmented samples (of the 20 created for each image) are shown on the right.

**Data Augmentation**

In each dataset, we applied different data augmentation techniques including: sequential rotation by 40 degrees, width shift with factor of 0.2, height shift with factor of 0.2, shear with a factor of 0.2, zooming with a range 0.2, horizontal flipping, and vertical flipping. Figure 5 shows some example images along with different augmented samples for the four different data classes. From Figure 5, it can be observed that noise has been added in some of the parts of the images. Therefore, we have also evaluated our method using only the center patch of the augmented samples. The downsampled and center patches are shown for two different input samples in Figure 6.

**Training Methodology**

In the first experiment, we trained with the IRRCNN architecture using the stochastic gradient descent (SGD) optimization function. We set the momentum to 0.9 and decay is calculated based on the initial learning rate and number epochs of the respective trial. We have experimented for three trials where 50 epochs are used in each trail. After 50 epochs, the learning rate is decreased by the factor of 10.

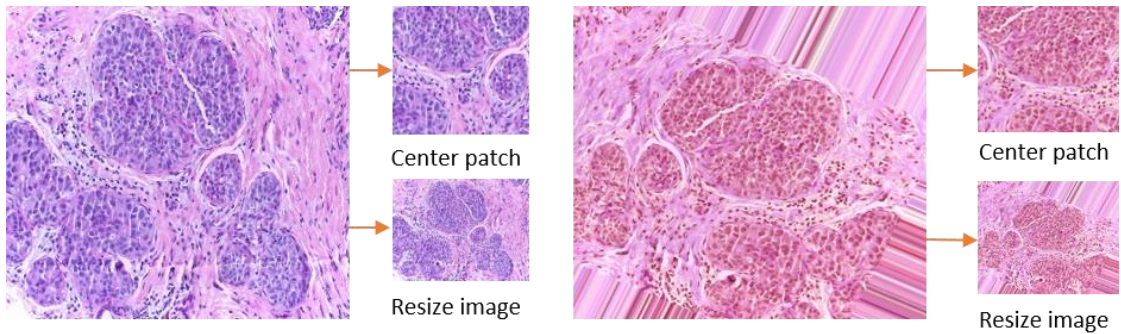

Figure 6. Center patch and resized images from an original sample (left) and from an augmented sample (right).

**Results and Discussion**

In this work, we introduced automated breast cancer classification for both the binary and multi-class problems on two different datasets. In the case of the multi-class BC classification problem, four and eight classes were considered in this implementation. We achieved state-of-the-art testing accuracy for both datasets.

**Results for BreakHis**

According to the work in [12,18], we considered two criteria on which to evaluate the performance of the IRRCNN model. We considered (1) image-level and (2) patient-level performance for multi-class classification for eight types of breast cancer that fall within the two main types (either benign or malignant). In addition, we have also evaluated the performance of a binary class system for the benign and malignant types. For image-level classification, we did not consider images with respect to patient. For this experiment, the images are organized into eight classes, and the images contain a magnification factor of either 40×, 100×, 200×, or 400×. Performance is measured with different evaluation metrics in this case. Two different performance criteria are considered to evaluate the performance of the IRRCNN deep learning approach as in [18]. First, we considered a patient-level performance analysis where the total number of patients is defined as $N_{np}$, the number of BC images of associated patient ($P$) is defined as $N_{ncp}$. The number of correctly classified images for a patient is denoted $N_{ntp}$. Equation (6) defines the patient score ($P_s$).

$$P_s = \frac{N_{ncp}}{N_{ntp}} \qquad (6)$$

The global patient recognition rate ($P_{rt}$) is defined in equation (7).

$$P_{rt} = \frac{\sum P_s}{N_{np}} \qquad (7)$$

We also calculated the performance of the IRRCNN approach for image-level classification. We define the total number of samples available for testing as $N_T$. The correctly classified number of histopathological samples is defined as $N_{CCT}$. The image-level recognition rate ($I_{rt}$) is expressed in equation (8).

$$I_{rt} = \frac{N_{CCT}}{N_T} \tag{8}$$

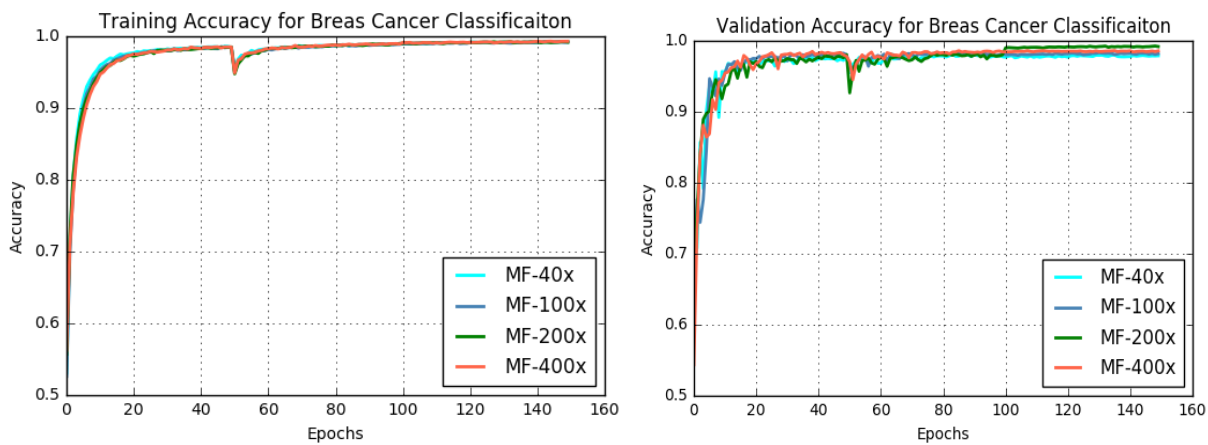

Figure 7. Training and validation accuracy for BC classification with 8 classes for the IRRCNN model at different magnification factors.

The training and validation accuracy of the IRRCNN model for breast cancer classification is shown in Figure 7. From this figure, it can be observed that the magnification factors of the samples have an impact on training and testing accuracy. We achieved the best training accuracy with a magnification factor of 100×, and the training accuracy achieved for data with a magnification factor of 200× is a very close second.

Table 3. Breast cancer classification results for multi-class (8 classes) using the BreakHis dataset.

|  | Methods | Year | Classification Rate (100R) at Magnification Factor | | | |
|---|---|---|---|---|---|---|
|  |  |  | 40× | 100× | 200× | 400× |
| Image Level | CNN+patches [15] | 2016 | 85.6 ± 4.8 | 83.5 ± 3.9 | 83.1 ± 1.9 | 80.8 ± 3.0 |
|  | LeNet + Aug [18] | 2017 | 40.1 ± 7.1 | 37.5 ± 6.7 | 40.1 ± 3.4 | 38.2 ± 5.9 |
|  | AlexNet + Aug [18] | 2017 | 70.1 ± 7.4 | 75.8 ± 5.4 | 73.6 ± 4.8 | 84.6 ± 1.8 |
|  | CSDCNN + Aug [18] | 2017 | 92.8 ± 2.1 | 93.9 ± 1.9 | 93.7 ± 2.2 | 92.9 ± 2.7 |
|  | IRRCNN +w/o Aug. | 2018 | 95.69 ± 1.18 | 95.37 ± 1.29 | 95.61 ± 1.37 | 95.15 ± 1.24 |
|  | IRRCNN + w Aug. | 2018 | 97.09 ± 1.06 | 97.57 ± 0.89 | 97.29 ± 1.09 | 97.22 ± 1.22 |
| Patient Level | LeNet + Aug [18] | 2017 | 48.2 ± 4.5 | 47.6 ± 7.5 | 45.5 ± 3.2 | 45.2 ± 8.2 |
|  | AlexNet + Aug [18] | 2017 | 74.6 ± 7.1 | 73.8 ± 4.5 | 76.4 ± 7.4 | 79.2 ± 7.6 |
|  | CSDCNN + Aug [18] | 2017 | 94.1 ± 2.1 | 93.2 ± 1.4 | 94.7 ± 3.6 | 93.5± 2.7 |
|  | IRRCNN +w/o Aug. | 2018 | 95.81 ± 1.81 | 94.44 ± 1.3 | 95.61 ± 2.9 | 94.32± 2.1 |
|  | IRRCNN + Aug. | 2018 | 96.76 ± 1.11 | 96.84±1.13 | 96.67±1.27 | 96.27±0.87 |

The testing accuracy for multi-class and binary BC classification is shown in Table 3 and Table 4 respectively. In both cases our IRRCNN based approaches show superior performance compared to existing DL based methods.

Table 4. Breast cancer classification results for binary classification (benign vs. malignant tumor) using the BreaKHis dataset.

| | Method | Year | Classification Rate at Magnification Factor | | | |
|---|---|---|---|---|---|---|
| | | | 40× | 100× | 200× | 400× |
| Image Level | CNN +fusion(sum, product, max) [15] (highest results) | 2016 | 85.6 ± 4.8 | 83.5 ± 3.9 | 83.6 ± 1.9 | 80.8 ± 3.0 |
| | AlexNet + Aug [18] | 2017 | 85.6 ± 4.8 | 83.5± 3.9 | 83.1 ± 1.9 | 80.8 ± 3.0 |
| | ASSVM [28] | | 94.97 | 93.62 | 94.54 | 94.42 |
| | CSDCNN + Aug [18] | 2017 | 95.80± 3.1 | 96.9 ± 1.9 | 96.7 ± 2.0 | 94.90 ± 2.8 |
| | IRRCNN | **2018** | **97.16 ± 1.37** | **96.84 ±1.34** | **96.61± 1.31** | **95.78 ± 1.44** |
| | IRRCNN + Aug | **2018** | **97.95± 1.07** | **97.57± 1.05** | **97.32± 1.22** | **97.36± 1.02** |
| Patient Level | CNN +fusion (sum, product, max) [15] | 2016 | 90.0 ± 6.7 | 88.4 ± 4.8 | 84.6 ± 4.2 | 86.10 ± 6.2 |
| | Bayramoglu et al. [14] | 2016 | 83.08 ± 2.08 | 83.17 ± 3.51 | 84.63 ± 2.72 | 82.10 ± 4.42 |
| | Multi-classifier by Gupta et al. [13] | 2017 | 87.2 ± 3.74 | 88.22 ± 3.23 | 88.89 ± 2.51 | 85.82 ± 3.81 |
| | CSDCNN + Aug [18] | 2017 | 92.8 ± 2.1 | 93.9 ± 1.9 | 93.7 ± 2.2 | 92.90 ± 2.7 |
| | IRRCNN +wo aug. | 2018 | 96.69 ±1.18 | 96.37 ±1.29 | 96.27±1 .57 | 96.15 ±1.61 |
| | IRRCNN + w. Aug. | 2018 | **97.60± 1.17** | **97.65± 1.20** | **97.56± 1.07** | **97.62± 1.13** |

In [15], the performance is analyzed with different fusion techniques including sum, product, and max. Thus, we compared against the highest accuracy reported in [15]. Our proposed method shows better performance in all cases.

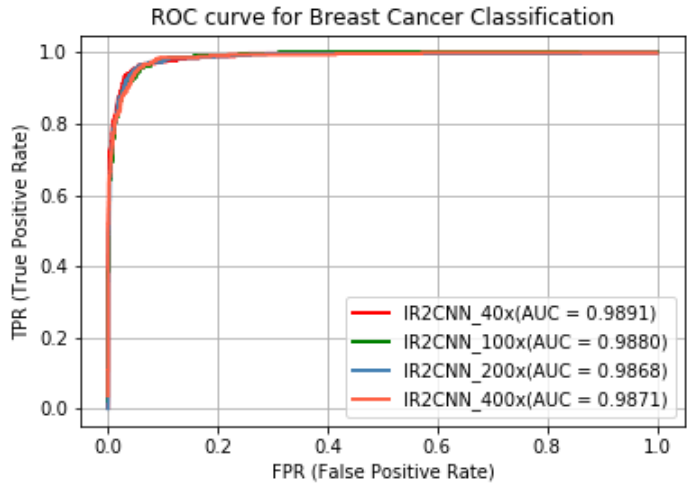

Figure 8. ROC curve with AUC for different magnification factors for eight class BC classification.

## Results for Breast Cancer Classification Challenge 2015

For the 2015 BC Classification Challenge dataset, the training and validation accuracy for different methods are shown in Figs. 9 (a) and (b) respectively. The experimental results when using resized and augmented samples show the highest training and validation accuracy according to Figure 9.

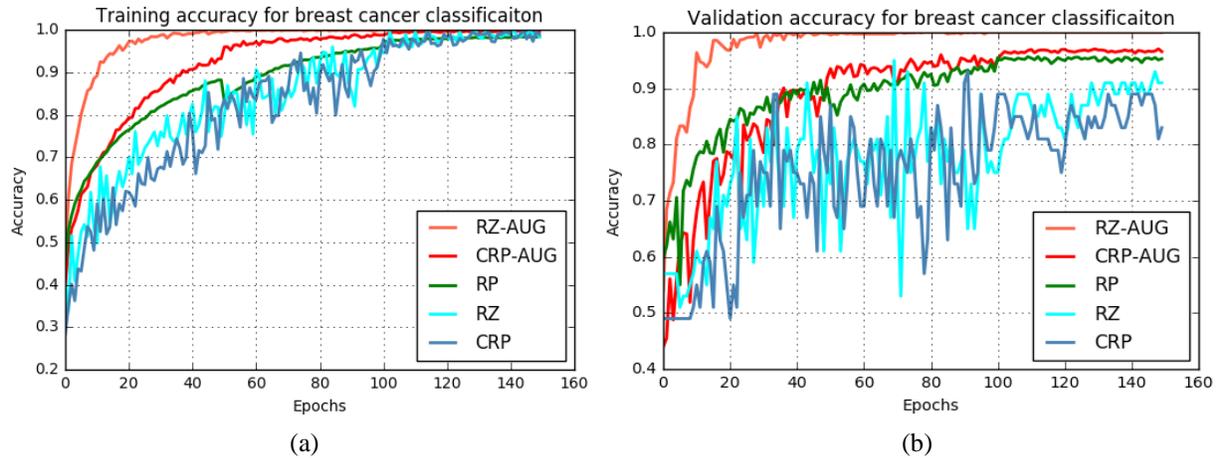

(a) (b)

Figure 9. Training and validation accuracy for the multi-class case using the 2015 BC Classification Challenge dataset. Sample set are either resized and augmented (RZ+AUG), center patch cropped and augmented (CRP+AUG), random patches (RP), sample resized (RZ), or center patch cropped (CRP).

**Patch-wise classification results:** The experimental results for different patch-based methods are shown in Tables 5 and 6. From the tables, for both binary and multi-class cases, the experiments with augmented center patches show the highest testing accuracy which is 97.51% and 97.11% respectively. Similar performance is observed with random patches, but the experiments with single center patches show the lowest accuracies which are 88.7% and 88.12% for the binary and multi-class cases respectively.

Table 5. Performance for center patches (CRP), augmented CRP, and random patches (RP) for the binary class case.

| CNN model | Methods | Year | Sensitivity | Specificity | Accuracy | AUC |
|---|---|---|---|---|---|---|
| CNN[17] | | | - | - | 0.776 | - |
| CNN+SVM[17] | | | - | - | 0.769 | - |
| IRRCNN | CRP | 2018 | 0.8732 | 0.8812 | 0.887 | 0.9239 |
| IRRCNN | CRP + Aug. | 2018 | 0.9452 | 0.9829 | 0.9751 | 0.9925 |
| IRRCNN | RP | 2018 | 0.9307 | 0.9797 | 0.9676 | 0.9882 |

Table 6. Performance for center patches (CRP), augmented CRP, and random patches (RP) for the multi-class case.

| CNN model | Criterions | Year | Sensitivity | Specificity | Accuracy | AUC |
|---|---|---|---|---|---|---|
| CNN[17] | | | - | - | 0.667 | - |
| CNN+SVM[17] | | | 0.810 | - | 0.650 | - |
| IRRCNN | CRP | 2018 | 0.868 | 0.8733 | 0.8812 | 0.9169 |
| IRRCNN | CRP + Aug. | 2018 | 0.9371 | 0.9809 | 0.9711 | 0.9905 |
| IRRCNN | RP | 2018 | 0.9290 | 0.9752 | 0.9634 | 0.9824 |

**Image-wise Classification results:** The experimental results for the binary and multi-class cases are given in Tables 7 and 8 respectively. We achieved 100% testing accuracy for the experiment with random patches using the WTA method. In addition, the experiments with augmented resized samples show 99.05% and 98.59% testing accuracy for the binary and multi-class cases respectively. The lowest testing accuracy was observed for single resized images.

Table 7. Performance for image-wise breast cancer classification for the binary case.

| CNN model | Criterion | Year | Sensitivity | Specificity | Accuracy | AUC |
|---|---|---|---|---|---|---|
| CNN[17] | | 2017 | - | - | 0.806 | - |
| CNN+SVM[17] | | 2017 | - | - | 0.833 | - |
| IRRCNN | RZ samples | 2018 | 0.878 | 0.926 | 0.884 | 0.912 |
| IRRCNN | RZ + Aug. | 2018 | 0.9831 | 0.9912 | 0.9905 | 0.9932 |
| IRRCNN | RP + WTA | 2018 | 1.00 | 1.00 | 1.00 | 1.00 |

Table 8. Performance for image-wise breast cancer classification for the multi-class case.

| CNN model | Criterion | Year | Sensitivity | Specificity | Accuracy | AUC |
|---|---|---|---|---|---|---|
| CNN[17] | | 2017 | - | - | 0.778 | - |
| CNN+SVM[17] | | 2017 | - | - | 0.778 | - |
| IRRCNN | RZ samples | 2018 | 0.889 | 0.916 | 0.9204 | 0.917 |
| IRRCNN | RZ + Aug. | 2018 | 0.9771 | 0.9889 | 0.9859 | 0.9905 |
| IRRCNN | RP + WTA | 2018 | 1.00 | 1.00 | 1.00 | 1.00 |

**Analysis and Comparison Against State-of-the-Art**

**Performance Analysis for the BreakHis Dataset:** Most previous studies have reported classification results for benign and malignant cases [8, 15, 18]. However, some studies have shown results for the multi-class problem for breast cancer classification [15,18]. These experiments have been conducted for both binary and multi-class problems on samples with magnification factors of 40×, 100×, 200×, and 400×. Based on the BreakHis dataset, different feature based approaches including PFTAS, GLCM, QDA, SVM, 1-NN, and RP were applied, and an accuracy of approximately 85% for patient level analysis was reported [8]. In addition, AlexNet was used for binary breast cancer recognition at different magnification factors, and the highest recognition accuracy achieved was 95.6±4.8% for image level analysis and 90.0±6.7% for patient level analysis [15]. Furthermore, the highest accuracies reported for classifying benign and malignant BC were 96.9±1.9% for the image level and 97.1±2.8% for the patient level [18]. For multi-class breast cancer classification, the best testing accuracies achieved were 93.9±1.9% and 94.7±3.9% for image level and patient level analysis respectively [18].

Alternatively, in this work we achieved 97.95±1.07% and 97.65±1.20% testing accuracy for benign and malignant BC classification for image and patient level analysis. Therefore, we have achieved a 1.05% and 0.55% improvement in average performance against the highest accuracies reported for image and patient level analysis in [18]. Furthermore, our proposed IRRCNN model produced testing accuracies of 97.57±0.89% and 96.84±1.13% for multi-class BC classification at the image level and patient level respectively. These results are a 3.67% and 2.14% improvement of average recognition accuracy compared to the latest reported performance [18].

**Performance Analysis for the 2015 BC Classification Challenge Dataset:** In 2014, Crus-Roa et al. proposed a CNN based method for classification with a patch based input, and they reported a sensitivity of 79.6% [29]. The highest accuracy that was reported in 2017 for four different types of breast cancers in the same dataset, and the experiments were conducted for both binary and multi-class breast cancer classification problems. As the data dimensionality is high (2040×1536 pixels), both image-level and patch-level analyses have been conducted for binary and multi-class breast cancer classification. A CNN approach was used, and the best results were reported for image-level classification which were 77.8% and 83.3% testing accuracy for four and two classes respectively [17]. On the contrary, we have conducted an experiment based on the IRRCNN model considering of different criteria including resizing, cropping, random patches, and different data augmentation techniques. For resized and augmented samples, we achieved 99.05% and 98.59% testing accuracy for binary and multi-class breast cancer recognition respectively. In addition, we achieved 100% testing performance for the experiment where the classification model is applied to random patches, followed by a winner take all method to produce the final results. Therefore, our method shows significant improvement in the state-of-the-art for both binary and multi-class breast cancer recognition on the 2015 BC Classification Challenge dataset. The computation times for these experiments are given in Table 9.

Table 9. Computational time per sample for the BC classification experiments.

| Dataset | Method | Total Time (s) | Number of Samples | Time per Sample (s) |
|---|---|---|---|---|
| BreakHis | Image Based | 72.06 | 8732 | 0.08 |
| BCC dataset 2015 | Image Based | 45.72 | 50 | 0.9144 |
| | Patch Based | 75.97 | 8742 | 0.008 |

## Conclusion

In this paper, we proposed binary and multi-class breast cancer recognition methods using the Inception Recurrent Residual Convolutional Neural Network (IRRCNN) model. The experiments were conducted using the IRRCNN model on two different benchmark datasets including BreakHis, and the 2015 Breast Cancer Classification Challenge, and performance was evaluated using different performance metrics. The performance of the proposed method was evaluated via image level, patient level, image based, and patch based analysis. We considered different criteria such as magnification factor, resized sample inputs, augmented patches and samples, and patch based classification in this implementation. The proposed approach shows approximately 3.67% and 2.14% improvement of average recognition accuracy on the BreakHis dataset against all results published in scientific reports as of 2016. In addition, this method shows 99.05% and 98.59% testing accuracy for binary and multi-class breast cancer recognition on the 2015 Breast Cancer Classification Challenge dataset, which is significantly higher than that of any other CNN based approach for image based and patch based recognition performance respectively. We have also evaluated the performance of the proposed method with random patches and Winner Take All (WTA) approaches for image based recognition and achieved 100% testing accuracy. Thus, the experimental results show state-of-the-art testing accuracy for breast cancer recognition compared against existing methods for both datasets.